\documentclass[letterpaper, 10 pt, conference]{ieeeconf}  

\IEEEoverridecommandlockouts                              

\usepackage{cite}
\usepackage{hyperref} 
\usepackage{graphicx}
\usepackage{multirow}
\usepackage{float}
\usepackage{siunitx}

\title{\LARGE \bf
PLATO: Planning with LLMs and Affordances for Tool Manipulation}
\author{Arvind Car$^1$, Sai Sravan Yarlagadda$^1$, Alison Bartsch$^1$, Abraham George$^1$ and Amir Barati Farimani$^1$
\thanks{$^1$A. Car, S. S. Yarlagadda, A. Bartsch, A. George and A. B. Farimani are with the Department of Mechanical Engineering at Carnegie Mellon University, Pittsburgh, PA, 15213, USA (e-mail: {\tt\small acar@andrew.cmu.edu}, {\tt\small barati@cmu.edu})}
}

\begin{document}
\maketitle
\thispagestyle{empty}
\pagestyle{empty}

\begin{abstract}
As robotic systems become increasingly integrated into complex real-world environments, there is a growing need for approaches that enable robots to understand and act upon natural language instructions without relying on extensive pre-programmed knowledge of their surroundings. This paper presents PLATO, an innovative system that addresses this challenge by leveraging specialized large language model agents to process natural language inputs, understand the environment, predict tool affordances, and generate executable actions for robotic systems. Unlike traditional systems that depend on hard-coded environmental information, PLATO employs a modular architecture of specialized agents to operate without any initial knowledge of the environment. These agents identify objects and their locations within the scene, generate a comprehensive high-level plan, translate this plan into a series of low-level actions, and verify the completion of each step. The system is particularly tested on challenging tool-use tasks, which involve handling diverse objects and require long-horizon planning. PLATO’s design allows it to adapt to dynamic and unstructured settings, significantly enhancing its flexibility and robustness. By evaluating the system across various complex scenarios, we demonstrate its capability to tackle a diverse range of tasks and offer a novel solution to integrate LLMs with robotic platforms, advancing the state-of-the-art in autonomous robotic task execution. For videos and prompt details, please see our project website: \href{https://sites.google.com/andrew.cmu.edu/plato}{https://sites.google.com/andrew.cmu.edu/plato}.
\end{abstract}

\section{INTRODUCTION}
\label{introduction}






As robotic technology continues to advance, the integration of robots into dynamic real-world environments remains a formidable challenge \cite{wong2018autonomous, kormushev2013reinforcement, ciocarlie2014towards, edsinger2006manipulation, billard2019trends}. Although the potential for robotic automation spans a variety of domains, from manufacturing \cite{liu2022robot} to service industries \cite{wang2022challenges}, current systems face significant limitations \cite{fernandez2021challenges}. 
Complex manipulation tasks often require substantial real-world datasets \cite{bartsch2024sculptdiff, bartsch2024sculptbot, george2024visuo}, and even approaches aiming to reduce the burden of data still lack in generalizability and adaptability outside of the training domain \cite{george2023minimizing, george2023one}. Recent works have shown that Large Language Models (LLMs) can provide world-knowledge, reducing the burden of data-collection \cite{ahn2022can}. However, many existing solutions that leverage this technology rely on extensive predefined knowledge and scripted routines, constraining their flexibility and adaptability in unpredictable settings and in scenarios that require long-horizon planning \cite{liang2023code, dikshit2023robochop}. These constraints hinder their effectiveness in real-world applications, where variables and environmental conditions can change rapidly and without warning. These limitations are further reflected in the autonomous tool use domains \cite{tang2024robotic, fang2020learning, qin2020keto}, as much of the works  improving manipulation generalizability assume simple pick place tasks to avoid addressing complex robot-object or robot-tool-object interactions \cite{ahn2022can, ding2023task, wu2023tidybot}. Even works that do investigate generalizability in tool handling make simplifying assumptions about being provided quality grasp locations to avoid reasoning about affordances and task-oriented grasp points \cite{xu2023creative}.

\begin{figure}
    \centering
    \includegraphics[width=\linewidth]{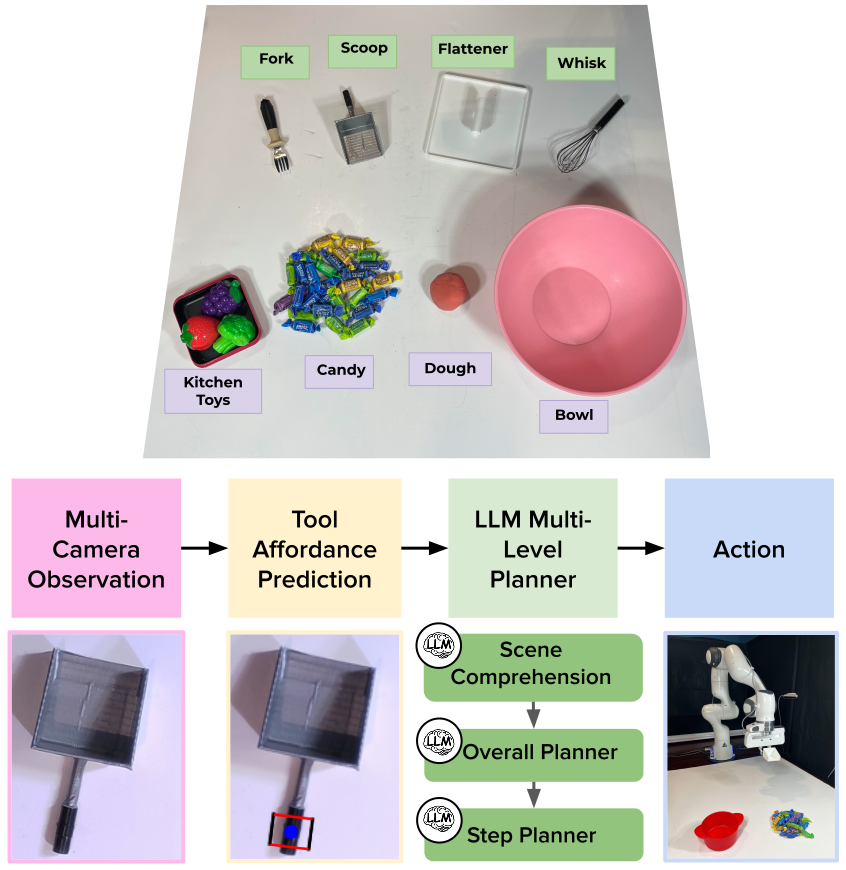}
    \vspace{-10pt}
    \caption{\textbf{PLATO Overview.} Our proposed system, PLATO, takes in an environmental observation, predicts the tool's affordance given the prompted task, and prompts a modular LLM framework to generate an action plan.}
    \label{fig:highlight}
    \vspace{-5pt}
\end{figure}

To address these challenges, we introduce PLATO, a framework for \textbf{P}lanning with \textbf{L}LMs and \textbf{A}ffordances for \textbf{To}ol manipulation. PLATO empowers robots with the ability to perform complex tool-based tasks autonomously across a diverse set of tools and objects. Unlike previous systems, PLATO operates without requiring prior knowledge of the environment \cite{xu2023creative}, enabling it to adapt dynamically to evolving conditions. This approach enhances the versatility of robotic systems, making them more practical and scalable across various domains. By focusing on generalizability and adaptability, our framework paves the way for more robust and flexible robotic applications, capable of meeting the demands of dynamic and unpredictable real-world scenarios. The key contributions of this work are as follows:
\begin{itemize}
    \item Introduced an architecture that utilizes a language model as both a high-level planner and a step planner. For instance, in cooking, the high-level planner enumerates all the steps from start to finish, while the step planner translates these high-level actions into low-level actions.
  \item Developed a grasping mechanism that utilizes a database of objects to identify the most similar object and infer appropriate grasping points for the target object.
  \item Developed a comprehensive end-to-end pipeline capable of adapting to diverse environments and executing tasks based on user queries.
\end{itemize}

\section{RELATED WORK}
\subsection{Large Language Models for Robotics}
The integration of Large Language Models (LLMs) into robotic applications for high-level planning marks a significant advancement, offering more natural and flexible human-robot interactions. However, much of the current work in this area utilizes a hierarchical planning approach that relies heavily on predefined information \cite{xu2023creative, bartsch2024llm}. These systems often invoke human-defined primitives \cite{ahn2022can, ding2023task}, or specialized motion planners \cite{chen2024autotamp, zhu2023pour}, to execute the plans generated by the LLMs, with details like object types, environment layouts, and task sequences being hard-coded. This reliance on predefined information limits the generalizability of the systems. While they may excel in specific tasks or environments, their ability to adapt to new or varying scenarios is constrained by the rigid structure of the encoded knowledge.

There have been substantial works addressing these limitations, with methods incorporating learned object and scene descriptors for improved generalizability \cite{wu2023tidybot}, explicitly including object semantic descriptions and captioning for adaptable navigation \cite{dorbala2023can}, with purely programmatic-based LLM planners which generate functions for the robot for simple pick and place tasks \cite{singh2023progprompt, liang2023code}, by grounding the LLM's behavior with 3D scene graphs \cite{rana2023sayplan}, or by incorporating diffusion models with LLMs for sub-goal generation \cite{sharan2024plan}. While these approaches have expanded the generalizability of LLM-based methods in terms of handling environmental variations, they remain focused on very simple manipulation actions, where various assumption of the robot-object interaction are made. As a result, the full potential of LLMs in creating adaptable and versatile robotic systems is not realized. To unlock this potential, there is a need for more dynamic and adaptive planning strategies that minimize dependency on hard-coded information or drastically simplifying assumptions, allowing for better generalization across a wider range of tasks, objects and environments.








\subsection{Robotic Tool Use}

The ability to use tools enables robots to perform tasks that would otherwise be beyond their capabilities. Tool manipulation has been a topic of interest in robotics for a long time, with numerous studies exploring how robots can use tools to complete specific tasks, such as screwing \cite{tang2024robotic}, hammering \cite{fang2020learning}, or reaching \cite{qin2020keto}. Although these approaches successfully generate actions for predefined tools, they fall short when it comes to handling novel objects or tasks. Some methods have employed tool randomization strategies to build robustness to variations within similar tool classes, by training keypoint detectors in simulation on generated "hammers" and "non-hammers" to identify grasp points, function points, effect points, and environment points \cite{qin2020keto}, or similarly by directly training a task-oriented grasp model in simulation with diverse generated variations of hammer-style tools \cite{fang2020learning}. However, these methods require specific tool variant training data and generally lack improvisational skills, as they do not account for the causal effects of tools or their interactions with other objects. Alternative methods have learned tool manipulation policies directly by predicting the flow of dense point clouds from tool manipulation demonstrations \cite{seita2023toolflownet}, or by updating tool manipulation control based on tactile observations \cite{shirai2023tactile}. These methods still require substantial tool and task-specific training data, making it difficult to generalize or adapt to tool and/or task variations. In contrast to these approaches, our proposed method harnesses the extensive prior knowledge and advanced planning capabilities of LLMs. By leveraging LLMs' world knowledge of object affordances, our approach enables robots to solve physical puzzles through creative reasoning and complex planning, offering solutions that go beyond predefined tool use scenarios.

\section{METHOD}
\begin{figure*}[h]
    \centering
    \includegraphics[width=0.95\linewidth]{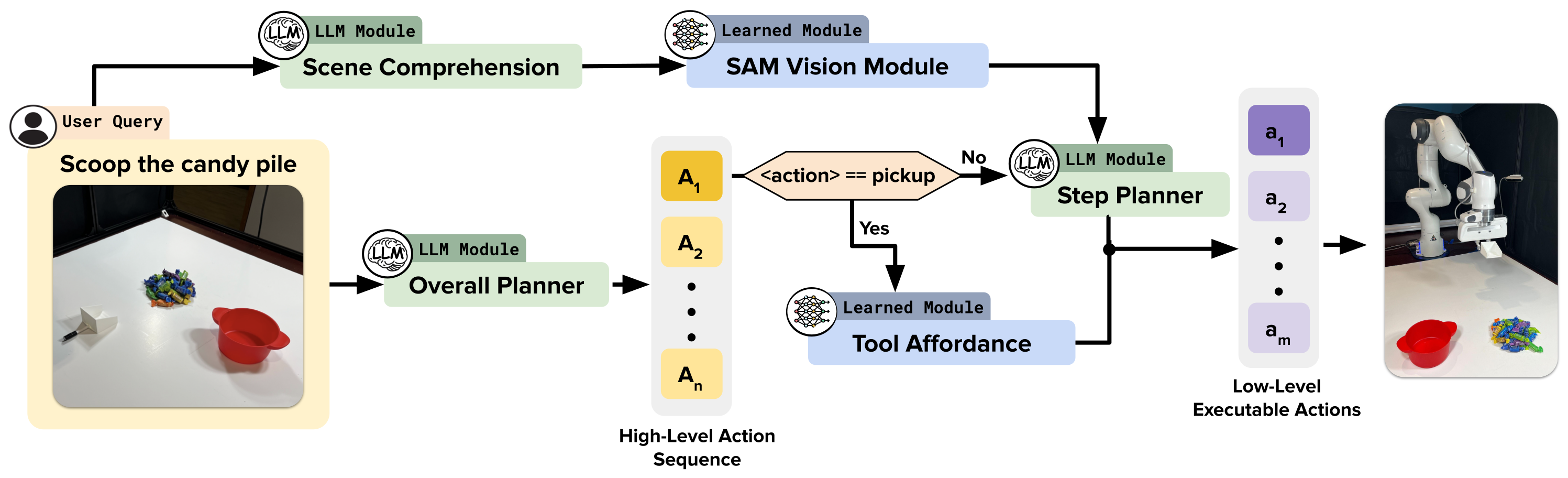}
    \caption{\textbf{Method Pipeline:} PLATO takes in a user prompt and multi-view images of the scene. These are passed to the Scene Comprehension LLM to list out the relevant objects present in the scene, and to classify them as a tool or not tool. This list of objects is passed to the SAM vision module, which segments the point clouds of each object, thereby getting its centroid and dimensions. Meanwhile, the Overall Planner LLM outputs a high-level sequence of commands, which are iteratively converted to low-level actions by the Step Planner LLM. These actions are sequentially executed by the robot.}
    \label{fig:pipeline}
\end{figure*}

In this work, we present PLATO, an agentic LLM framework which takes in a user prompt containing a task to be executed, breaks it down into smaller actionable chunks and executes them with no prior information about the environment. The framework consists of four major modules, each playing a crucial role in the overall functionality and adaptability of the system, with the LLM modules powered by GPT-4o \cite{achiam2023gpt}. A schematic of the overall pipeline is visualized in Figure \ref{fig:pipeline}. We will provide detailed explanations for each of the key modules in the following subsections. For prompt details, please see our project website.

\subsection{Scene Comprehension}
This module is responsible for analyzing a scene in the workspace and identifying the various objects present that are relevant to the task that needs to be performed. The model receives the user query, alongside the image of the scene. The model is prompted to observe the image, and generate all the objects that are present in the scene relevant to the User Query, as well as a binary value denoting whether that object is a ``tool" or ``not tool". This classification is used for further downstream tasks that may require the object to be grasped. This initial comprehension step is crucial for providing accurate and relevant input to the subsequent modules, ensuring that the robotic system has a clear understanding of the scene it will be interacting with.

\subsection{High level Planner}
The next step in our framework is the high-level planner. This module is responsible for breaking down the User Query into a series of high-level steps ($A_1, A_2, ... A_n)$ required to complete the User Query. The input to this module is the User Query and the list of objects generated by the Scene Comprehension module. The output of the module is structured as a series of steps, with each step being represented by 4 key phrases, namely $<$action$>$, $<$location$>$, $<$object$>$,  and $<$tool$>$. This representation is used to create clear and concise abstractions of the high level planning steps, which can be easily understood by subsequent modules. The $<$action$>$ represents the overall task being executed by the parallel plate gripper (Eg: pickup, place, scoop, flatten, etc.). The $<$location$>$ is a semantic description that represents where the action is to be performed, and is defined based on the current position of an object, original position of an object, or the robot home pose. The $<$object$>$ represents what the action is being performed on and is selected from the list provided by the Scene Comprehension module. The $<$tool$>$ represents the object that the gripper holds while performing an action. Either/both $<$object$>$ and $<$tool$>$ can take `none' values. For example, if the Overall Planner Module was given the User Query of ``Make a salad", its first step ($A_1$) might be [`pickup', `original position of tomato', `tomato', `none']. Here, $<$tool$>$ is `none' because at the start of the action; the gripper is not holding any object. 

\subsection{Vision Module}
\label{segmentation}
The Vision Module takes in the images of the scene, captured by 4 RGBD cameras and outputs the centroid and dimensions of all the objects listed by the Scene Comprehension Module. This allows us to ground the actions produced by the Planning modules with real world environmental information. Apart from enabling the localization of objects during execution, it also provides the Planner modules with richer information about the objects they interact with, thus guiding them to make more effective and informed plans. The following steps outline the process through which the Vision Module extracts and processes the necessary information from the scene. First, we utilized an existing framework, Grounding Dino \cite{liu2023grounding}, which takes both a color image and a semantic input specifying the object of interest. Grounding Dino identifies and localizes objects within the image captured by each camera by generating bounding boxes around them and assigning them a confidence score. Among the set of bounding boxes produced, we select the ones with a confidence score above a certain fixed threshold. After obtaining the bounding boxes, we generate masks to segment the object within each bounding box \cite{kirillov2023segment}. We apply the same masks to the corresponding depth images as well. Using the masked depth and color images, we can generate a partial point cloud for each object mask, and use precomputed transformation matrices to convert them to the robot coordinate frame. We also keep track of the logit score ($l_{ij}$) associated with each object mask ($m_{ij}$). We denote these partial pointclouds by $pc_{ij}$ where i is the camera number (i = 1, ...4) and j is the mask number for a given camera. In order to map the pointclouds of a given object across different cameras views, we use distance based thresholding. This allows us to combine these partial pointclouds ($pc_{1p}$, $pc_{2q}$, $pc_{3r}$, $pc_{4s}$) to get a complete pointcloud ($PC_{k})$ of an object. We also aggregate the logit scores associated with each of the partial point clouds, to get the total logit score $L_k$. For a given semantic query, we obtain a list of point clouds along with their corresponding logit scores. The point cloud with the highest $L_k$ is selected as the queried object. This approach not only provides a more accurate 3D representation of the object but also enhances object segmentation accuracy compared to single-view SAM by effectively addressing partial occlusion issues. Once we have the complete pointcloud for an object, we denoise it, and obtain dimensional information regarding the object (centroid ($x,y,z$), length ($L$), width ($W$), height ($H$)).

\begin{figure*}
  \centering
  \includegraphics[width=0.78\linewidth]{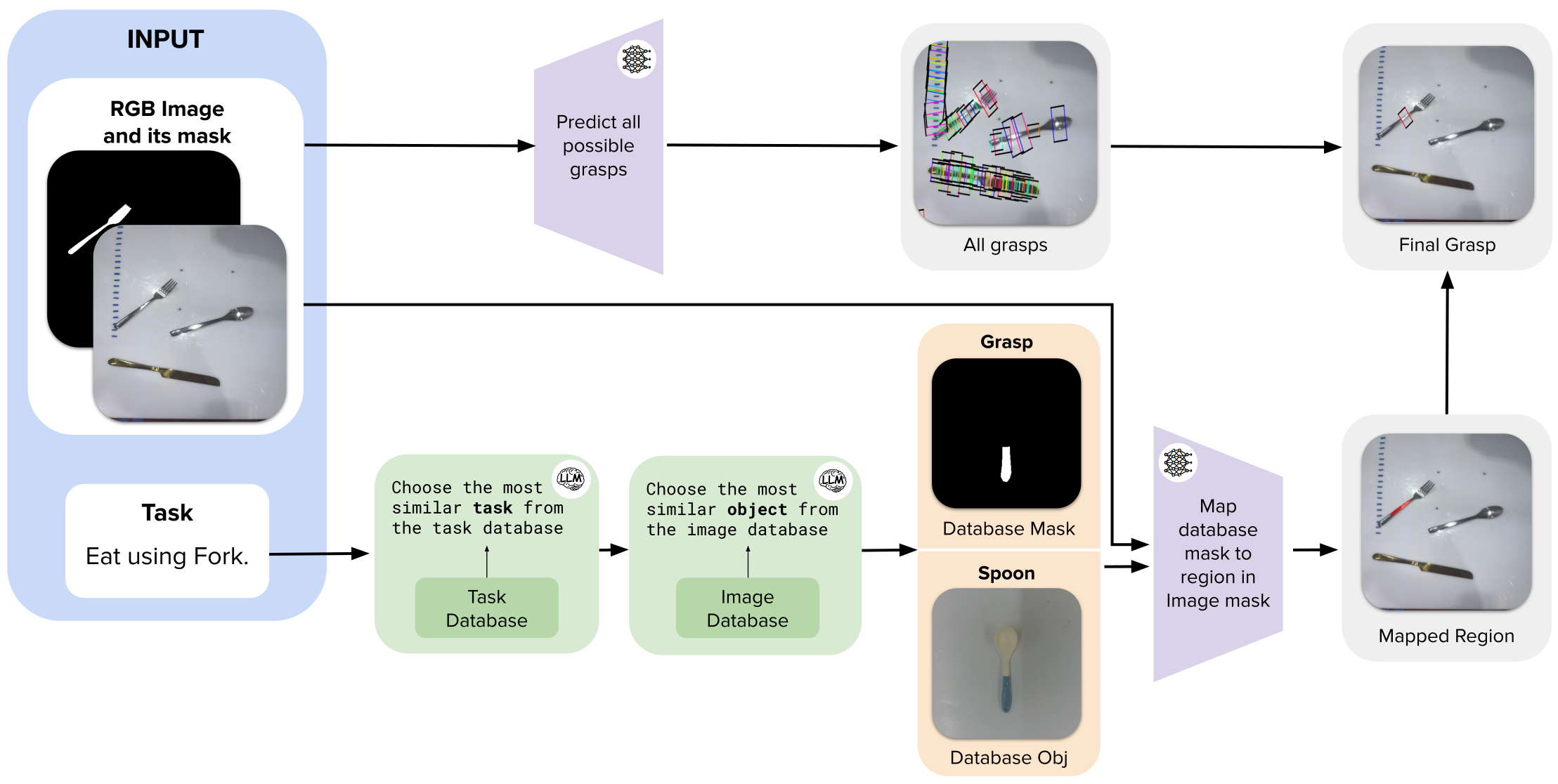}
  \caption{\label{fig:main} \textbf{Task-Oriented Grasping Module.} The process starts by generating a mask of the target object using an overhead image captured by a camera mounted on the end-effector. Potential grasps are then generated and refined based on the tool’s graspable region. The task and tool are passed to an LLM, which maps the query tool to the most similar one in the affordance model's database. The model uses this mapping to refine grasp selection, and if the mapping fails, the mask of the entire tool is used to guide the grasp choice. }
  \label{figure-tog}
\end{figure*}

\subsection{Step Planner}
The Step Planner is a crucial component that takes the output from the High-Level Planner and converts it into low-level actions executable by a parallel plate gripper. This module ensures that each high-level instruction is broken down into precise, actionable steps, leveraging a combination of go-to poses, gripper states, and tilt commands. The Step Planner bridges the gap between human-understandable commands and machine-executable actions, enabling the robotic system to perform tasks accurately and efficiently. The Step Planner is used only when the step provided by the Overall Planner is not a task that involves picking up a tool (defined as anything with a handle). For non-pickup tasks, and pickup tasks that don't involve tools, we use the Step Planner module instead. For all other kinds of steps ($A_t$), the step planner decomposes it to a series of robot actions ($a_1, a_2, ... a_m)$, each consisting of either a $'Go-to:\ <$location$> + \ (\Delta x, \Delta y, \Delta z)'$ command, $'Grasp:\ <$0/1$>'$ to open or close the gripper or $'Tilt:(\phi, \theta, \psi)'$ command. The Step Planner is given the coordinates and dimensions of the object and/or tool involved in $A_t$, in order to better guide its plan. Additionally, to prevent any redundant actions from being performed, the step planner is also provided with the last executed step $A_{t-1}$. For instance, a high-level command such as [`place', `original position of table', `table', `hammer'] is transformed into specific steps like 1. `Go-to: original position of table + (0, 0, 2) cm', 2. `Grasp: 0', and 3. `Go-to: original position of table + (0, 0, 10) cm'. The dimensional information provided to the the Step Planner allows to provide accurate deltas when describing Go-to commands This methodical approach ensures that each action is executed with precision, taking into account the robot's coordinate system and the physical constraints of the task. The step planner's ability to translate high-level instructions into detailed, executable steps is essential for the successful operation of the robotic framework in dynamic and unstructured environments.

\subsection{Task Oriented Grasping}
Upon locating the centroid of the object the gripper will interact with, the framework first determines whether the respective object is classified as a tool that must be grasped in a particular way, or an object that can be grasped anywhere to perform the task. In the case of the latter, the framework simply grasps the object around the centroid and moves on to the next step. But in the case of the former, we utilise an affordance model that is capable of identifying the graspable points based on the query task and tool. For our affordance model, we augment an existing one-shot affordance model with task based reasoning \cite{os_tog}. We expand this framework's generalizability by utilizing the inherent tool-grasping reasoning of LLMs to map the query tools to the tools in the affordance model's database. This allows us to utilize tools not previously seen by the model, thus making this  a zero-shot effort. 

This framework comprises of the following steps. Firstly, the model generates a mask \cite{kirillov2023segment} of the target object using an overhead image taken by the end-effector mounted camera, positioned over the centroid of the tool (the coordinates of which is obtained from the SAM Vision Module). Secondly, the model generates all potential grasps and their corresponding scores for each object visible in the overhead image. These grasps will then be narrowed down in further modules, based on the tool's graspable region. Next, the query task (in this case, ``pickup") and the query tool are passed to an LLM agent, tasked with mapping the query object to the most similar tool (in terms of grasp-ability) in the model's database. The model then attempts to map the mask of the graspable region of the database tool to the image of the query tool. If the model succeeds in mapping the mask of the database tool to the query image, the generated mask is used to shortlist the relevant grasps generated previously, from which the grasp with the highest score is chosen. As a fail-safe, if the model fails in the mask mapping process, the model uses the mask of the whole tool, instead of just the graspable region, to shortlist the previously generated grasps. Finally, once the optimal grasp is identified on a pixel-level, the coordinates for the grasp are obtained using the depth camera information for the scene to identify the coordinates of the grasp region.


\begin{figure}
    \centering
    \vspace{-10pt}
    \includegraphics[width=1.0\linewidth]{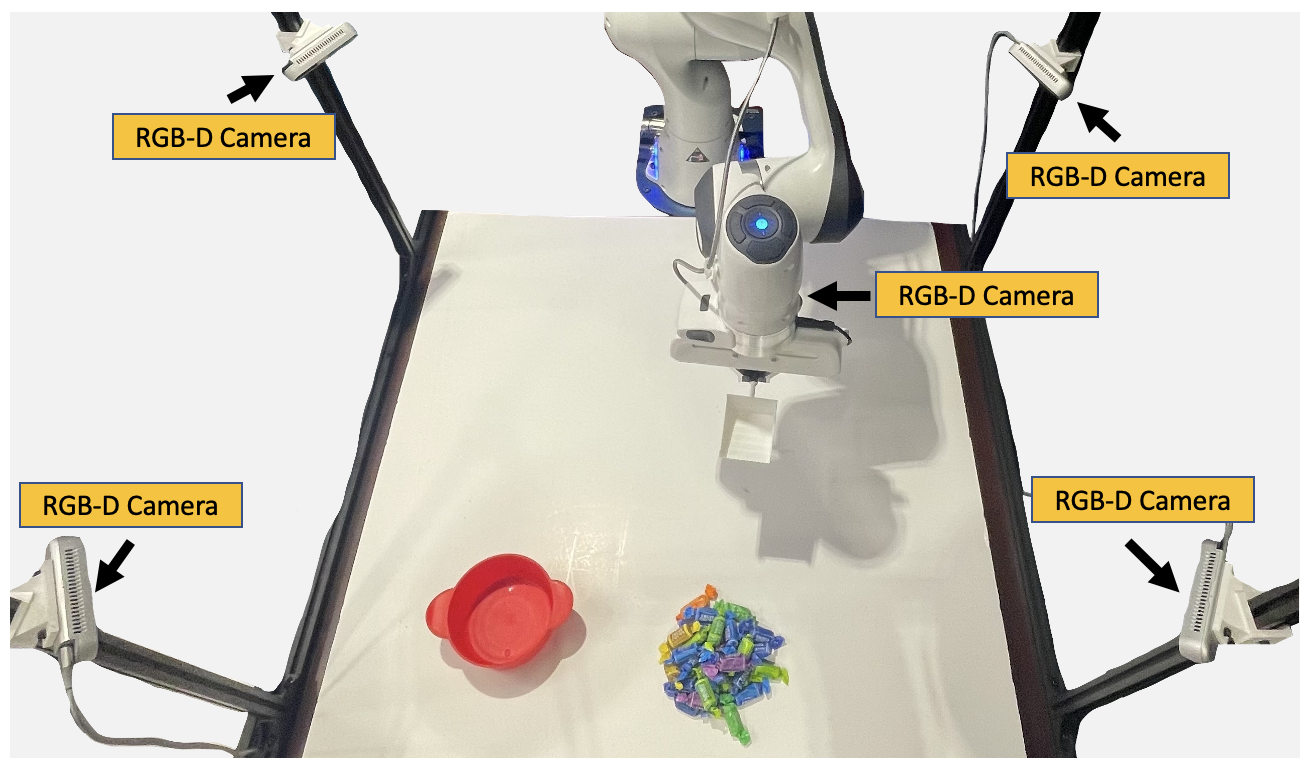}
    \caption{\textbf{Hardware Setup.} A visualization of the 7DoF Franka robot in the table-top manipulation workspace with 4 Intel RealSense D415 cameras to observe the scene, and a single wrist-mounted Intel RealSense D415 camera.}
    \label{fig:setup}
\end{figure}

\section{Physical Setup}
\label{setup}

As mentioned in Section \ref{segmentation}, we construct a comprehensive 3D point cloud of the entire environment to facilitate accurate scene understanding and task execution. This 3D point cloud is generated using 5 Intel RealSense D415 cameras. These cameras are mounted on a custom-built cage that surrounds the scene to capture a the environment from multiple angles. Out of the 5 cameras, 4 of the cameras are positioned on the camera cage to cover different perspectives of the workspace, while the fifth camera is mounted directly on the parallel plate gripper of the Franka Emika Robot (7 Degrees of Freedom). The placement of the fifth camera on the gripper is solely to identify task oriented grasps. A detailed view of the setup is shown in the Figure \ref{fig:setup}.

\section{Experiments}

To assess the performance of our system, we conducted three categories of experiments, with each category consisting of three individual tasks (combinations of objects and goals), with 10 trials each. The first category, Single-Task Grasping, involved straightforward pick-and-place tasks that did not require querying the affordance model or reasoning about tool use. The goal of these experiments was to evaluate the overall architecture, with a particular focus on assessing the performance of the vision pipeline across two basic pick-and-place type tasks. The second category, Single-Task Tool Use, was designed to test the framework's ability to reason about tool use and evaluate the affordance model. This was examined through three experiments involving different types of tools and use cases. Finally, we conducted Multi-Task Tool Use experiments to evaluate the system's performance on a sequence of long-horizon tasks, which included multiple tool-object interactions.
In order to assess the outcome of each experiment, we established partial success metrics to evaluate progress at various stages of the process. The metrics vary across the three types of experiments, based on what the experiment is trying to evaluate. The results of these experiments are shown in Table \ref{tab:results}, and videos can be found on the project website.

\subsection{Single-Task Grasping} 
The two single task experiments conducted were ``\textit{Place $<$obj1$>$ next to $<$obj2$>$}" and ``\textit{Place $<$obj1$>$ within $<$obj2$>$}". The objects were chosen from a toy kitchen set, comprising small-scale plastic food items and containers. Since we wanted to test the ability of the vision module to identify and localize different objects, we varied the object types and positions within the 10 runs for each of the two experiments. 25\% success corresponds to the scene comprehension agent correctly listing out all the objects in the scene. 50\% corresponds to the Vision module correctly identifying the relevant objects and their locations. 75\% corresponds to $<$obj1$>$ being picked correctly. 100\% corresponds to $<$obj1$>$ being placed correctly, relative to $<$obj2$>$.

\subsection{Single-Task Tool Use} 
Three experiments were conducted to assess the model's performance on Single-Task Tool use tasks: ``Scoop up candy," ``Flatten the ball of dough," and ``Whisk the empty bowl." These experiments were designed to evaluate both the affordance model's capabilities and the Step Planner module's ability to break down high-level instructions such as ``scoop," ``flatten," and ``whisk" into basic low-level robot actions (go-to poses, tilt angles, and gripper positions). In these experiments, a 25\% success rate indicates the scene comprehension agent correctly identifies all objects in the scene. A 50\% success rate signifies that the Vision module accurately detects the relevant objects and their locations. A 75\% success rate reflects the affordance model predicting the appropriate grasps (i.e., the tool is held in a way that it can be used for the task). Full success, or 100\%, indicates that the Step Planner has performed the necessary movements accurately.

\subsection{Multi-Task Tool Use} 
Three experiments were conducted to evaluate the model's performance on Multi-Task Tool use tasks. The first experiment, with a single tool and multiple objects, was ``Scoop the candy and place it inside a bowl". The second task, with a single object and multiple tools, was ``Flatten the dough and poke holes in it". The third task, involving multiple tools and multiple objects, was ``Flatten the dough, and scoop candy onto it". While the tools and objects remained consistent for a given experiment, their positions were varied across different runs of the same experiment. In these experiments, 25\% success corresponds to the scene comprehension agent correctly listing out all the objects in the scene. 50\% success corresponds to the Vision module correctly identifying the relevant objects and their locations. 75\% success corresponds to the first tool-object interaction being performed correctly. 100\% success corresponds to the second object tool interaction being performed correctly.


\section{Results}

The most common failure mode was the Step Planner's difficulty in understanding the state/orientation of the grasped tool. Providing only object dimensions and locations is insufficient for reasoning about orientations and interactions. A richer feedback mechanism describing the scene's state could improve this. Passing scene images didn’t resolve the issue, as tool-specific details were still missed.
The system performs well on simple pick-and-place tasks and translating well-defined tasks (e.g., ``scoop," ``flatten") into robot actions, but struggles with ambiguous tasks like ``whisk." The LLM agent in the Tool Affordance model excelled at mapping query tools to database tools, but expanding the tool database would enhance performance, as it currently relies heavily on the overall object mask. The vision module (SAM) also had trouble distinguishing similar tools (e.g., spatula vs. scoop), leading to lower success rates in Multi-Task Tool experiments.

An ablation study, shown in Table \ref{tab:ablation} verified the importance of the affordance model. In this study, instead of using the affordance model, tools were grasped at the centroid, as with objects, which drastically reduced success rates for Scooping and Whisking, since they require specific grasps.  This is because they involve tools that need to be grasped in a particular way, in order for them to be used as intended. However, in the case of the Flattening task, there is actually a reduction in the grasping related errors, as the tool's handle is very near the centroid, and attempting to grasp it, even with the wrong orientation, rotates the tool to make the grasp successful. This resulted in no grasping related errors for this task.

\begin{table}[]
\caption{\textbf{Quantitative Results.} We present the success rate for different phases of each task. STG is single-task grasping, STT is single-task tool use, and MTT is multi-task tool use.}\label{tab:results} 
\centering
\begin{tabular}{@{\extracolsep{\fill}}lllllll}
        \hline
        \textbf{} & \textbf{} & \textbf{25\scriptsize \%} & \textbf{50\scriptsize \%} & \textbf{75\scriptsize \%} & \textbf{100\scriptsize \%} \\ 
        \hline
        \hline
        \multirow{2}{*}{\textbf{STG}} & \scriptsize Place next to & 10 & 9  & 8 & 7 \\
          & \scriptsize Place inside & 10 & 8 & 7 & 6 \\
        \hline
        
        \multirow{3}{*}{\textbf{STT}} & \scriptsize Scoop candy & 10 & 8 & 8 & 5 \\
          & \scriptsize Flatten dough & 10 & 10 & 7 & 6 \\
          & \scriptsize Whisking motion & 10 & 9 & 8 & 4 \\
        \hline
        
        \multirow{3}{*}{\textbf{MTT}} & \scriptsize Flatten dough \& poke holes & 10 & 9  & 6 & 3 \\
          & \scriptsize Scoop candy \& pour into bowl & 10 & 6 & 5 & 3 \\
          & \scriptsize Scoop candy onto flattened dough & 10 & 1 & 0 & 0 \\
        \hline
    \end{tabular}
\end{table}

\begin{table}[]
\caption{\textbf{Ablation Results.} We present the success rate for different phases of each single-task tool use comparing our method to the affordance-free baseline.}\label{tab:ablation} 
\centering
\begin{tabular}{@{\extracolsep{\fill}}lllllll}
        \hline
        \textbf{} & \textbf{} & \textbf{25\%} & \textbf{50\%} & \textbf{75\%} & \textbf{100\%} \\ 
        \hline
        \hline
        \multirow{3}{*}{\textbf{PLATO (ours)}} & \scriptsize Scoop candy & 10 & 8 & 8 & 5 \\
          & \scriptsize Flatten dough & 10 & 10 & 7 & 6 \\
          & \scriptsize Whisking motion & 10 & 9 & 8 & 4 \\
        \hline
        
        \multirow{3}{*}{\textbf{No Affordances}} & \scriptsize Scoop candy & 10 & 7 & 0 & 0 \\
          & \scriptsize Flatten dough & 10 & 8 & 8 & 6 \\
          & \scriptsize Whisking motion & 10 & 9 & 4 & 1 \\
        \hline

    \end{tabular}
\end{table}

\subsection{Single-Task Grasping}

The framework performed well on simple pick-and-place tasks, with failures primarily arising from the vision module's ability to accurately mask objects in the scene. This issue could be mitigated by having SAM generate multiple masks, then using an LLM agent to select the most appropriate mask for the given object query, rather than simply relying on the highest-scoring mask generated by SAM. The performance of the scene comprehension module suggests that this approach could improve the accuracy of the object masks produced by the vision module. When the masking was done accurately, the framework effectively handled the spatial reasoning required and successfully executed the tasks most of the time.

\subsection{Single-Task Tool Use}
Apart from the object-segmentation-related errors seen in the previous set of tasks, the main source of errors for the Single Task tool use experiments was the step planner module's ability to convert high-level actions into low-level executable robot commands. Although, the Step planner module was able to decompose the more well-defined actions like ``Scoop" (hold the tool in-line with the base of the object -$>$ push towards the object -$>$ tilt upwards to secure the scooped object) and ``Flatten" (hold the tool above the object -$>$ attempt to go below the object), it struggled with the more complex motion of ``whisk", which it tried to perform as a series of oscillatory go-to poses, or rotations. This is also partly due to a lack of feedback regarding the tool and object poses as it is performing these actions.

\subsection{Multi-Task Tool Use}
While the framework's performance remained stable with the addition of more objects, it declined significantly when multiple tools were involved, especially if they were structurally similar. This issue was particularly evident in the ``Flatten the dough and scoop candy onto it" task, which used the scoop and flattener tools. The Vision module (specifically SAM) struggled to differentiate between these two tools, resulting in a high number of failed experiments as the robot would attempt to do a sub-task with the wrong tool.

\section{Conclusion}
In this work, we present PLATO, a novel agentic LLM framework that successfully decomposes complex tasks from user prompts into smaller, actionable steps and executes them autonomously without prior hard-coded knowledge of the environment. Our framework gathers information from the environment using an image segmentation module and utilizes depth data to accurately localize objects. Our experiments showcase the strength of leveraging GPT-4o-powered LLM modules to achieve generalization and adaptability across diverse tasks. Beyond action planning, the affordance model also utilizes an LLM agent to map queried tools to corresponding database tools, enhancing tool-specific interactions. Our experiments also highlight the need for a more robust, semantically informed object segmentation module and a richer feedback system for improved performance.


\bibliographystyle{IEEEtran}
\bibliography{references}


\end{document}